\documentclass[times, review,10pt]{elsarticle}
\usepackage[title]{appendix}%
\usepackage{amssymb}
\usepackage{amsmath}
\usepackage{yfonts}
\usepackage[colorinlistoftodos]{todonotes}
\usepackage{url}

\usepackage{setspace}

\usepackage{hyperref}
\journal{}

\begin{document}
\doublespacing

\begin{frontmatter}



\title{FP-THD: Full page transcription of historical documents}
%Text recognition of Historical Documents with few labeled data
% Joint layout analysis, masked autoencoder and vision transformer for the transcription of historical documents in Latin

%\title{Joint Layout Analysis, Masked Autoencoder and Vision Transformer for civil law Historical Document Digitization}

%\title{An extension of masked autoencoder for the full processing of multipage historical documents}

\author{Hala Neji\corref{cor1}\fnref{label1}}
%\author[label1]{Hala Neji}
\author[label1]{Javier Nogueras-Iso}
%\ead{jnog@unizar.es}
\author[label1]{Javier Lacasta}
%\ead{jlacasta@unizar.es}
\author[label1]{Miguel Ángel Latre}
%\ead{latre@unizar.es}
\author[label2]{Francisco Javier García-Marco}
%\ead{jgarcia@unizar.es}
\cortext[cor1]{Corresponding author. E-mail address: hala.neji@unizar.es}



\affiliation[label1]{organization={Aragon Institute of Engineering Research (I3A),Universidad de Zaragoza },%Department and Organization
            addressline={María de Luna 1}, 
            city={Zaragoza},
            postcode={50018}, 
            state={Aragón},
            country={Spain}}

\affiliation[label2]{organization={Institute of Heritage and Humanities (IPH),Universidad de Zaragoza },%Department and Organization
            addressline={Pedro Cerbuna 12}, 
            city={Zaragoza},
            postcode={50009}, 
            state={Aragón},
            country={Spain}}

%% Abstract
\begin{abstract}

The transcription of historical documents written in Latin in XV and XVI centuries has special challenges as it must maintain the characters and special symbols that have distinct meanings to ensure that historical texts retain their original style and significance. This work proposes a pipeline for the transcription of historical documents preserving these special features. We propose to extend an existing text line recognition method with a layout analysis model. We analyze historical text images using a layout analysis model to extract text lines, which are then processed by an OCR model to generate a fully digitized page. We showed that our pipeline facilitates the processing of the page and produces an efficient result. We evaluated our approach on multiple datasets and demonstrate that the masked autoencoder effectively processes different types of text, including handwritten, printed and multi-language.

%We achieve competitive results using only 134 training pages.
%Moreover, it tested on the transcription of a doctrinal work written by Miguel del Molino, a well-known civil law expert in XV century in the Aragon kingdom. We achieve character error recognition of 1.7\% and word error recognition of 4.5\% without expanded abbreviations.
%Additionally, our algorithm achieved a Character Error Rate (CER) of  1.3\% on Rodrigo dataset  and 4.46 \%  on Benthman dataset, demonstrating strong performance, particularly on Rodrigo dataset.
% our algorithm tested on the Rodrigo and Benthman dataset performance improves by 1.3\%  and 4.46 \% character error recognition  
%\todo[inline]{The following sentences should be revised after having a full version of the paper. Perhaps, they are too detailed for an abstract.}

\end{abstract}


\begin{highlights}
\item Proposal of pipeline for full page image transcription specialised in medieval Latin.
\item Integration and customization of models for layout analysis and OCR.
\item Evaluation with existing datasets and new dataset for Latin transcription.
\item Effective performance for printed and handwritten transcription.
\end{highlights}

%% Keywords

\end{frontmatter}



\section{Introduction}
\label{sec1}

Apparently, the applicability of Optical Character Recognition (OCR) algorithms to digitised documents seems a simple process. First, layout analysis \cite{santos2025comparative} must be performed to identify different regions of the page, such as columns, paragraphs, text lines and images. Second, the system then identifies and recognises lines of text, ensuring the correct reading order. Then, a text recognition tool is used to convert the image into digital text. Once recognised, the text areas are properly organised to preserve the original structure. Finally, the system generates the extracted text, making it ready for digital storage.

However, the recognition of handwritten and printed text in historical documents continues to be a persistent challenge due to several problem. On the one hand, the conditions of the hard-copy document due to the degradation of the paper \cite{zhou2023review} or the ink stains, among of other problems. In those cases, noise reduction \cite{neji2019adversarial}, image denoising \cite{neji2024doc} , deblurring \cite{neji2024deep} or binarization \cite{hamza2025rseunet} must be applied to remove distortion, improve contrast \cite{ziran2025enhancing} and make the text clearer. Sometimes, it is necessary to repair the damaged document images to improve the readability of text \cite{yang2025predicting}.

On the other hand, the majority of modern transcription tools either do not identify historical characters or convert them into modern ones, altering the original text. In many cases, the origin of this wrong recognition of characters by modern OCR tools is the lack of ground truth data, but in historical documents old characters and special symbols often have unique meanings that should be preserved. For instance, symbols such as \textfrak{s} employed for a long ``s'' are converted either wrongly into ``f'', or converted into the modern ``s''; and ligatures (æ, œ) are converted into ``ae'', ``oe''.
%; many other characters that are converted to modern characters. 

%\todo[inline]{los dos siguientes parrafos no los veo separados de este. Son una caracterización de la no alineación entre símbolos usados antes y los de ahora. Al separarlo, parecen ideas distintas.}

%\todo[inline]{Yo le daría la vuelta al párrafo. El párrafo anterior describes un problema y pones ejemplos, en este lo haces al revés, pones un ejemplo y luego describes el problema. Yo lo haría como en el anterior. hablar del problema de las abreviaciones y comentar el ejemplo de la tilde.}

%h1

To preserve the original old characters of historical documents, we need an advanced OCR system to customize the recognition and preservation of special characters, ensuring that historical texts retain their original style and meaning. This is particularly important for historical research, where details such as ligatures, archaic spellings or handwritten annotations provide a valuable context. In medieval Latin manuscripts, the tilde (˜) functioned as a form of scribal shorthand, commonly indicating omitted letters or syllables: especially nasal consonants like \textit{m} and \textit{n}, or frequent grammatical endings. These abbreviations were widely used to save space and speed up writing, with forms such as \textit{``n\={o}''} for \textit{non}, \textit{``d\~{n}i''} for \textit{domini}, \textit{``\~{e}im''} for \textit{enim}, `\textit{`d\~{e}''} for \textit{dem}, \textit{``\~{q}''}  for \textit{quod}, or \textit{``\~{e}''} representing variations like \textit{en}, \textit{em}, or even \textit{in}, depending on the context \cite{spina2023artificial, aguilar2025dual}.

%To ensure that the digital version remains true to the original,

The purpose of this paper is to develop our own transcription pipeline that retains all characters exactly as they appear in the image.
%The idea for our pipeline came from the lack of tools to preserve the authenticity of old pages and the need to detect their various features. 
The idea for our pipeline came from the need to detect the features of pages in historical documents to preserve their authenticity. 
%Our goal is to develop a pipeline that simplifies the testing of existing models with images and addresses these challenges.
Before applying the OCR, we found that analyzing the layout is a crucial step in text image recognition. To predict the sequence of characters in a text image, it is easier to first extract text lines and classify the page content before performing OCR.
Therefore, we extend a masked autoencoder \cite{li2025htr} for printed and handwritten text recognition for the full processing of multi-page historical documents. 
This extension is built on top of a model originally designed to recognize text in lines that is modified to develop a system taking a whole page image as input and producing the transcription of the whole page. First, the system parses the page with a  layout analysis model called ParseNet \cite{kodym2021page}, which examines a document to identify and understand the structure. This first step detect the regions of text lines and crop them into text line images. During the second step a masked autoencoder with vision transformer processes the line images to recognized the text in each line. Finally, the system sorts the lines and regions, and then reconstruct the pages. It generates an XML file that contains the layout baseline and a text file that contains the full page recognition. 

The remainder of this paper is structured as follows. Section \ref{sec:relatedWork} presents an overview of the existing works related to text recognition, highlighting the different strategies used to extract text from images. Section \ref{sec:method} describes the main components in the design of our proposed pipeline for the transcription of the content of a full page. Then, Section \ref{sec:experiments} presents the experimental setup and results. Finally, the paper ends with discussion of the findings (Section \ref{sec:discussion}) and some concluding remarks (Section \ref{sec:conclusions}).

\section{Related work}
\label{sec:relatedWork}

The task of extracting text from document images combines several key areas of research. The related work in this field depends on the type of text and the machine learning methods used to extract it. The main goal is to extract text from an image. Whereas many approaches start with a layout analysis step to detect and organize different text regions to facilitate the transcription, others use end-to-end systems to extract text directly. There are also other works focused on text line transcription as input of the model. 

%We organize the related work into three main categories: layout detection, which focuses on identifying the structural elements of a document such as paragraphs, text baselines, tables, and headers; optical character recognition (OCR), which converts the visual text content into machine-readable format; and integrated solutions, which aim to combine layout understanding and OCR into end-to-end systems for comprehensive document analysis. Below, we review representative works from each of these domains.

%\todo[inline]{me resulta rara esta categorización del estado del arte cuando solo hay un ejemplo en las categorías 2 y 3.}

In the context of layout analysis challenge, as a first step to extract the region of text, Gr{\"u}ning et al. \cite{gruning2019two} presented an effective two-stage method for text line detection in historical documents using ARU-Net. By leveraging deep learning and clustering techniques, this approach achieves high accuracy while handling complex layouts and varying text orientations on cBad competitions \cite{diem2017cbad}. Kodym et al. \cite{kodym2021page} extended the latter model and presented a Convolutional Neural Network (CNN) model, called ParseNet, which was designed for the simultaneous detection of text baselines, text line polygons, and text blocks of printed and handwritten documents. The flexible design of this model can process complex document layouts, making it a powerful tool for digitizing historical texts with minimal preprocessing.
As a complementary work to this layout analysis model, Koh{\'u}t et al. \cite{kohut2021ts} implemented an OCR model, called TS-Net, that has been trained to detect archaic characters (e.g., \textfrak{s}) and switch between text transcription styles. However, it requires labeled data for each style, which can be hard to obtain. If the style switch fails, the output can be inconsistent. In addition, TS-Net focuses more on style detection than on handling real-world challenges like poor image quality or diverse handwriting.

 % Joint layout analysis, character detection and recognition for historical document digitization     ### detection recognition joint both of them 

In addition, many techniques are proposed for text line segmentation and
baseline detection techniques for historical document analysis \cite{rabaev2025recent}. Lee et al. \cite{lee2024algorithm} proposed a post-OCR algorithm for line segmentation and reading order sorting based on adjacent character detection to improve the digitization quality of Chinese historical texts.

 In the context of joint layout analysis, character detection and recognition,
 Ma et al. \cite{ma2020joint} proposed an end-to-end trainable framework for historical document restoration that uses two parallel branches: one for character localization and recognition, and another for layout analysis to accurately restore content while maintaining the correct reading order. Preceding the parallel network, a Feature Pyramid Network processes the features  extracted by ResNet-50 \cite{he2016deep}. 
 %The character branch performs character classification and bounding box regression, while the layout branch uses a fully convolutional network to generate a binary mask of text regions. A line detection method is then applied to this mask to identify text lines. In the post-processing stage, characters are grouped into columns and text lines based on the combined results of both branches. Finally, the document content is reconstructed and output in the correct reading order. Some additional steps are applied only during inference.
Van et al. \cite{van2024loghi} also proposed  a framework called Loghi for joint layout analysis and text recognition of historical documents. They integrate modular tools for layout analysis, text line detection, OCR, and language detection, achieving high accuracy on both handwritten and printed archives. Although Loghi offers a complete pipeline for processing historical documents, it has certain limitations. The system is heavily dependent on document layout analysis, which can fail in highly degraded or irregular manuscripts. It also uses predefined components for tasks like text recognition and segmentation, which limits its flexibility and adaptability across different document types. Additionally, Loghi may not perform as well on complex handwritten content or languages not well represented in its training data.

Focusing on isolated OCR tasks, Shiferaw et al. \cite{shiferaw2025handwritten} proposed a hybrid approach to handwritten Amharic character recognition by combining CNNs with machine learning classifiers. While effective for isolated character recognition, this approach exhibits key limitations when applied to full OCR pipelines. First, the method focuses mainly on recognizing individual characters, which makes it less effective for full words or lines of text, where understanding the spacing and relationships between characters is important. Second, because the feature extraction and classification steps are trained separately, the system can not fully learn from the data in one unified process. This limits its ability to adapt to different layouts, noise, or variations in handwriting. Last, the use of pre-trained CNNs that were not designed for handwritten text can lead to features that do not work well for historical or more complex documents.

Li et al. \cite{li2025htr} also developed a model for handwritten text recognition of text lines extending a Masked Autoencoder (MAE) introduced by He et al. \cite{he2022masked} and focused on self-supervised pretraining for vision tasks, where a portion of the image patches is randomly masked, and the model learns to reconstruct the missing information. The original method also employs Vision Transformers (ViTs) to learn robust image representations for tasks like classification and object detection. The handwritten text recognition model proposed by Li et al. \cite{li2025htr} adapts the ViT architecture specifically for recognizing handwritten text. It incorporates a CNN feature extractor to capture rich spatial features from the image before passing the tokens to the transformer encoder. Unlike MAE's random patch masking, this model employs a span masking strategy, which is more suitable for the sequential nature of text. Moreover, instead of using reconstruction loss like MAE \cite{he2022masked}, the handwritten text recognition model is trained with CTC (Connectionist Temporal Classification) loss to predict character sequences. Additionally, the model removes the class token and utilizes sinusoidal position embeddings, reflecting task-specific modifications to the ViT architecture. Therefore, the main differences are in the task, the masking method, and the design choices for text recognition.

Among recent approach in handwritten character recognition, Imboden et al. \cite{campana2025open} proposed an open-source workflow. Their system integrates different open-source tools for preprocessing, feature extraction, and machine learning, and it is particularly effective when applied to historical documents. This approach contributes to the field by providing a flexible and reproducible framework that supports research in digital humanities and the preservation of written heritage. However, their method was not tested on large datasets and depends a lot on good image preprocessing, which can reduce accuracy on damaged documents.

Taking into account the limitations identified in previous works, such as the lack of end-to-end learning, limited adaptability to historical handwriting, and the use of modernized character transcriptions, the aim of this work is to propose a pipeline that not only transcribes historical handwritten text with high accuracy, but also respects the original character forms and visual features of the documents, ensuring faithful preservation and interpretation. For this purpose we propose a solution that integrates the  model proposed by Kodym et al. \cite{kodym2021page} for layout analysis as an initial step, which is followed by an OCR model extending the masked autoencoder proposed by Li et al. \cite{li2025htr}.

%highlight the need for a more suitable approach. Many existing tools fail to preserve the original appearance and form of historical characters, which is crucial for accurate digital representation and scholarly use. Our work is motivated by the need to fill this gap. We . 

\section{Method}
\label{sec:method}

\begin{figure}[h]
\center
%\setlength{\fboxrule}{0.2pt}
 %\fbox{
 \includegraphics[width=\textwidth]{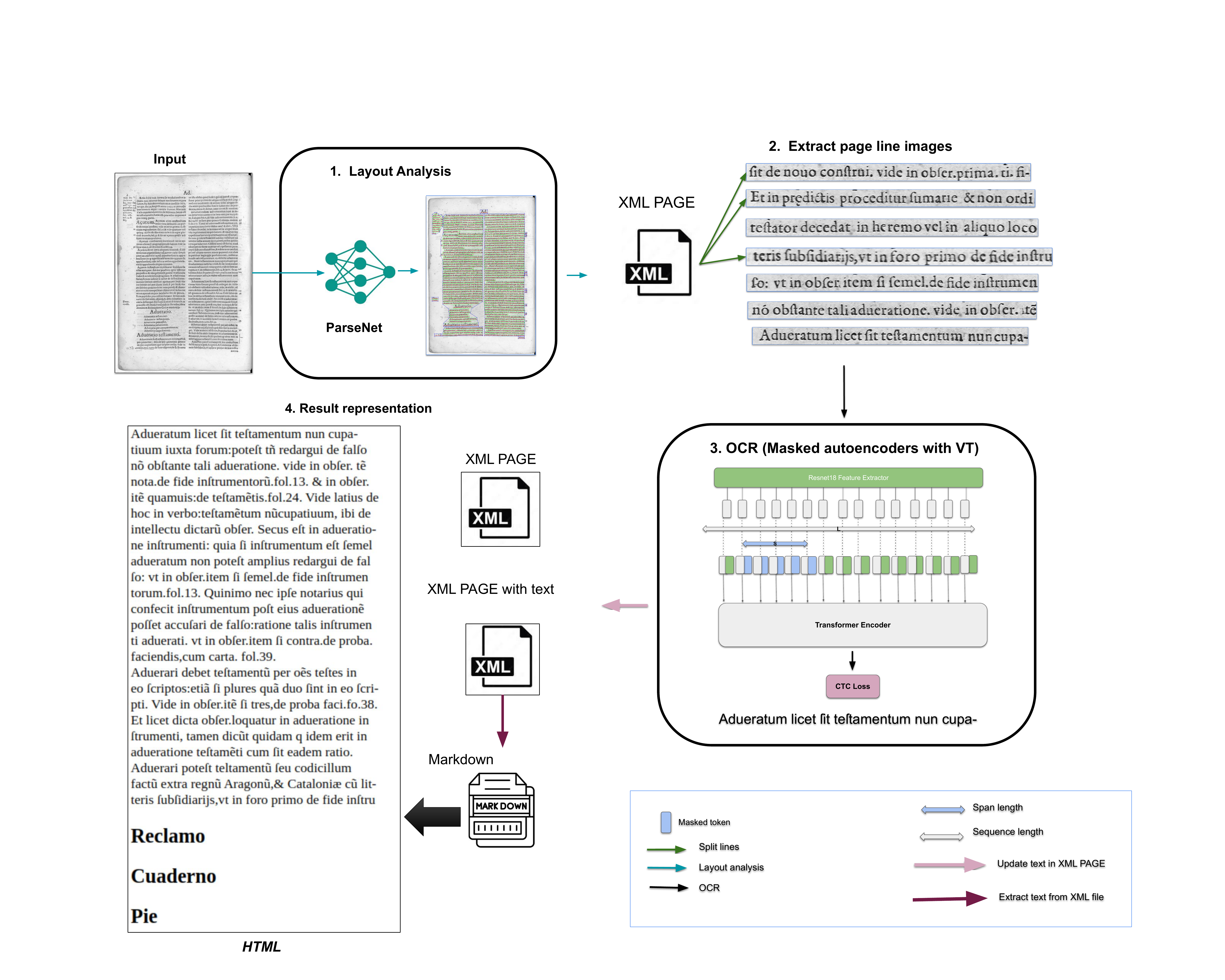}
\caption{FP-THD architecture Overview: Layout Analysis and Masked Auto-encoder with Vision Transformer }
\label{fig:architectureWorkflow}
\end{figure}

Figure \ref{fig:architectureWorkflow} shows the architecture of our proposed pipeline system, which consists of four components: layout analysis, extraction of line images, OCR, and result representation. The division into components is based on the assumption that transcription is easier if we classify first the page content and extract text lines before applying OCR.
 The following subsections describe the insights of these four components.

%\todo[inline]{The figure should be provided as PDF to have a better resolution} 

\subsection{Layout analysis}
\label{sec:layout}
%To predict the sequence of characters in a text image, it is easier to first extract text lines and classify the page content before applying OCR. This step-by-step approach improves accuracy by ensuring that OCR is applied to well-segmented and properly structured text regions.

The objective of this component is to segment and structure the content of the document based on its layout, i.e. identify the different areas in the document such as text blocks, headings and paragraphs, and determine the positions of the text lines ensuring the correct text flow. Due to the success of the layout analysis method proposed by Kodym et al. \cite{kodym2021page}, we selected ParseNet as the basis for our layout analysis component. ParseNet is a deep learning architecture designed for document layout and text line analysis. It uses an encoder–decoder convolutional network to detect baselines and estimate the vertical extent of each text line. The model produces probability maps for text baselines, ascenders, and descenders, which are then combined to precisely locate and extract lines. This approach allows the system to handle complex, curved, or skewed text lines and to generate geometrically normalized outputs suitable for OCR processing.
 
As shown in Figure~\ref{fig:architectureWorkflow}, the input of this component is a document page image containing the text to be transcribed. First, this page is processed by ParseNet, which generates an output image where each detected line is highlighted or marked to reflect the layout analysis.
Alongside this image, an XML output with the layout using the PAGE (Page Analysis and Ground-Truth Elements) format \cite{pletschacher2010page}. This XML format was defined by the University of Salford to link the text obtained by an OCR tool to specific areas of the digitized image. In a similar way to other formats like hOCR \cite{hOCRSpec} or ALTO \cite{ALTO2022}, this format has the advantage that there are various tools that allow the joint visualization of the image and the extracted text \cite{primaTools}.

\subsection{Extraction of page line images}
\label{sec:textLine}
As a result of the previous step, an XML file was generated for each digitized page in PAGE format. Using this XML file, a cropping algorithm extracts each text line region and passes the list of text line images as input to the OCR component of the pipeline. 

The cropping step extracts each detected text line from the full page image using baseline coordinates and height estimates provided by the layout analysis network. Each line is geometrically rectified along its predicted baseline and rescaled to a standardized height of 50 pixels, as set in the system configuration. This ensures all line images have uniform height despite the original text size or layout. By aligning curved or tilted lines and normalizing their height, the process produces clean, consistent line images. These images are then saved temporarily for OCR recognition, improving recognition accuracy by providing standardized input data.

\subsection{OCR}
\label{MAE}
Encouraged by the success of the Masked Autoencoder with Vision Transformer (MAE-ViT) model \cite{li2025htr}, we decided to apply it for our OCR component and extend it to recognize various text categories, including printed, handwritten, old, and modern texts. 

Several factors determine our choice of Masked autoencoder. On the one hand, it contains a CNN-based feature extractor which was used to derive features for each input token, allowing for robust feature extraction. A ResNet-18 \cite{he2016deep} is employed for the feature extraction. 

On the other hand, it includes a span feature masking technique that replaces masking tokens with learnable tokens, helping to reduce the risk of overfitting even in the case of having a corpus with a limited number of pages.
Additionally, it utilizes the Sharpness-Aware Minimization (SAM) optimizer to guide the model towards convergence at flatter minima. The span masking strategy serves as a form of regularization specially designed for handwritten text recognition. After the input image is processed by a CNN feature extractor, it produces a feature map that is flattened into a sequence of tokens (vectors), each representing part of the image features. Instead of masking isolated single tokens randomly, the ``span mask'' technique masks contiguous groups of tokens (spans) of fixed maximum length. Specifically, a certain portion of the tokens in the sequence is masked out in spans of up to 8 connected tokens. These masked spans are then replaced by learnable mask tokens, meaning the model learns a vector representation specifically for the masked input regions.
In short, it combines powerful CNN-based feature extraction and transformer based sequence modeling with thoughtful masking and training techniques to achieve highly accurate handwritten text recognition, even with limited training data. In addition, this method does not include any post-processing steps, and the model does not leverage a language model (LM) to correct the output of the OCR.

\subsection{Result representation}
\label{sec:reconstruct}

The pipeline generates several output files that capture and represent the OCR results at different stages. Initially, the layout processor produces an XML in PAGE format reflecting the original detected layout structure of the page, including regions and lines without recognized text. After the OCR processor recognizes text line by line, this information must be integrated back producing an updated XML file where each text line element contains the transcription from the OCR. For this purpose, this component uses the naming conventions previously applied to generate the file names of the input line images in order to locate the XML element corresponding to the correct region and line where the text content should be added as an XML subelement.

%%%%%%%%%%%%%%%%%%%%%%%%%%%%%%%%%%%%%%%%%%%%%%In the case of the dataset that contains full pages and without layout analysis information, we apply the full pipline. Which means, after updating the XML PAGE with the text lines recognized by our model, we generate two XML files: One has the layout, and the other the updated recognized text. The reconstruction of transcription of the full page is done only when we want to reconstruct all the pages.
%one contains the layout and the other containing the updated recognized text.

%In addition to the page layout, which defines the position of each text line, we update the layout with the model’s output by inserting the recognized text into its corresponding line position.

Apart from the PAGE format, we also wanted to generate a transcription representation that could be used by experts to revise the text and incorporate annotations. Initially, we considered the use of Text Encoding Initiative (TEI) \cite{TEI} for this representation. It is a XML-based format that allows specialized tags with a wide variety of semantics to be defined in order to annotate named entities or details of the transcription. However, editing an XML format remains a difficult task for transcription experts without the help of adapted tools. Therefore, instead of TEI we have enriched this component with the functionality of generating a representation in Markdown format \cite{markdown} as an initial step to facilitate the revision of text and the addition of small annotations (footnotes, links). Inspired from many existing conventions for marking up email messages using plain text, Markdown is a lightweight markup language that aims to achieve maximum readability and ease of publication. Besides, it has the advantage of being editable and viewable on collaborative work platforms such as GitHub, which adds support for version control as well.

Last, as an additional output, this component also generates a plain TXT file for each page with the all the text lines. This continuous textual document facilitates an easier text access and downstream processing. In addition, this allows the measurement of OCR performance (Test Loss, CER,WER) in different experiments through the comparison with the ground truth text.

% Additionally, a plain text file is generated that consolidates all recognized text lines into a  These outputs together enable comprehensive documentation and verification of both the layout detection and text recognition results.

%which involves generating a text file for each page, followed by conversion into Markdown format. 

%%%%%%%There is still some manual work involved, specifically in marking the beginning of each concept. This step is expected to be automated in future work.
%%%%%It is particularly important in this project, as the identification of the beginning of concepts in this book \cite{molino1585} help the experts that will revise and verify the latin transcription, given that some letters are unclear ever for humans. This process includes all pages of the book: that contains the concept name, the `folio' number (whether recto or verso), the text alignment (right or left), and the transcription of the text for all pages.

%%%%%%%This aligns with our deliverable requirements, as the final output was intended to be in HTML format. 

%%%%%%and the purpose of this project is to create a transcription of the Miguel del Molino book \cite{molino1585}   
%%%%%%%%% Moreover, the text extracted for each section has been uploaded to our platform, accessible at the following link: \href{https://migueldelmolino.es/}{Platform for the analysis of the documentary heritage of Aragonese regional law}.

\section{Experiments}
\label{sec:experiments}

This section describes the applicability of our proposed pipeline for the transcription of different types of text. First, section~\ref{sec:datasets} describes the different datasets that have been used in our experiments. Then, sections~\ref{sec:handwritten} and \ref{sec:printed} describe how the pipeline has been trained for the transcription of handwritten and printed texts. The code and data specifically created for this research is available at this repository: \url{https://github.com/IAAA-Lab/FP-THD}.

%\todo[inline]{We need to upload the code. Otherwise, our paper will be probably rejected, or we will have to fix this. You can create number folders for each step: 1) layout analysis, 2) extraction of lines, 3) OCR, 4) result presentation. In case we haven't modified anything for ParseNet or MaE-Vit, add pointers to the original repositories. But at least, it should be possible to add some scripts whenever we have adjusted something or those related to the reconstruction of pages, the generation of markdown, some scripts to generate metrics, ... We can also add a folder for datasets: rodrigo, bentham, molino. In the case of rodrigo and bentham, add pointers. In the case of Molino, add the data that we have prepared, and even the scripts to semi-automate their generation.}

%We evaluate our pipeline on different datasets, as described in the following subsection. The experimental results clearly demonstrate the effectiveness of our approach, achieving competitive performance compared to state-of-the-art methods.

\subsection{Datasets and environment configuration}
\label{sec:datasets}

In our experiments we have used: two existing datasets of handwritten texts, called Rodrigo and Bentham, and employed in the research literature of transcription systems; and one dataset of printed texts, called Molino, which was specifically prepared for this work.

\begin{figure}[h]
\center
\includegraphics[width=1.0\textwidth]{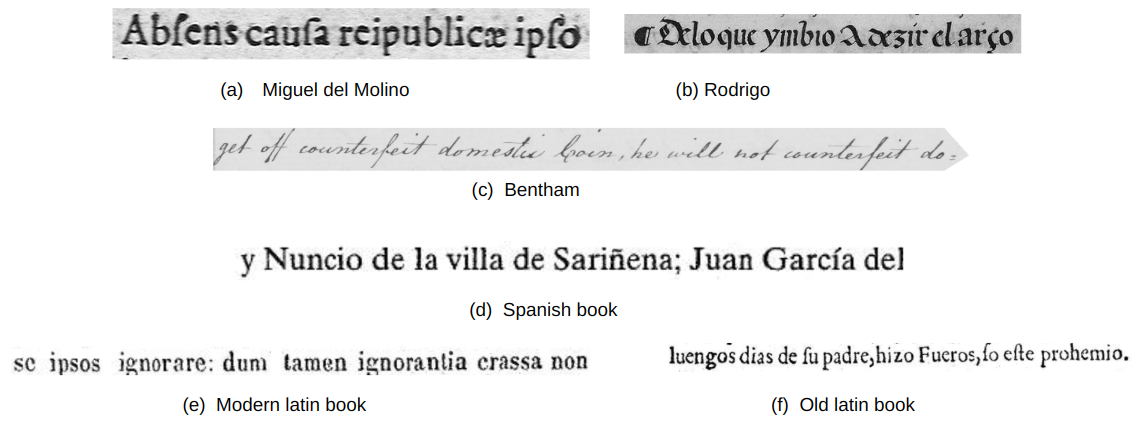}
\caption{ Example text lines by datasets. }
\label{fig:examplesPages}
\end{figure}

The Rodrigo dataset is a corpus obtained from the digitisation of a handwritten book in old Castilian with chronicles of the history of Spain authored by an archbishop called Rodrigo in 1545 \cite{emilio_granell_2018_1490009}. It contains 853 pages with Gothic typography, which are organised in 307 chapters. Figure~\ref{fig:examplesPages} (b) illustrates a representative text line sample from this dataset.

The Bentham dataset \cite{causer2012building} is a widely used benchmark in the field of handwritten text recognition, derived from the manuscripts of the English philosopher Jeremy Bentham (1748–1832) \cite{bentham}. It consists of high-resolution scans of historical handwritten documents, which are part of the Bentham Papers collection held by University College London. These manuscripts are characterized by a wide variety of writing styles, ink types, and page conditions, reflecting the natural challenges encountered in real-world historical document recognition tasks. Figure~\ref{fig:examplesPages} (c) illustrates a representative text line sample from this dataset. 

Due to the lack of available datasets of historical printed text in medieval Latin with abbreviations, we created our own dataset called Molino. This dataset is part of a project to create a platform for the analysis of historical documents of civil law in the historical kingdom of Aragon (Spain) \cite{neji2024first}. This corpus consists of 143 page images with Roman typograpy extracted from four sources: 23 pages extracted from a book edited in 1667 \cite{fueros1667} with civil law regulations written in medieval Latin with abbreviations and old Spanish (Figure~\ref{fig:examplesPages} (f) illustrates a representative text line sample); 69 pages of the same civil law regulations edited in 1866 and written in modern Latin without abbreviations and old Spanish (Figure \ref{fig:examplesPages} (e) illustrates a representative text line sample); 15 pages extracted from a book with the translation into Spanish of Latin fragments in the previous 2 sources \cite{delgado1991} (Figure~\ref{fig:examplesPages} (d) illustrates a representative text line sample); and 37 pages extracted from a doctrinal book explaining how to apply civil law regulations and authored by Miguel del Molino (a civil law expert) in an edition of 1585 using medieval Latin with abbreviations \cite{molino1585} (Figure~\ref{fig:examplesPages} (a) illustrates a representative text line sample).

%\subsubsection{Pero layout dataset}
%The PERO Layout Dataset \footnote{https://www.fit.vut.cz/person/ikodym/pero\_layout.zip} is a comprehensive dataset designed specifically for document layout analysis. It includes a diverse collection of scanned documents, both historical and modern, with detailed annotations of layout elements such as text blocks, paragraphs, headings, tables, images, and other structural components. This dataset is particularly useful for training and evaluating machine learning models for tasks like document segmentation, region classification, and automated document structure recognition. 

\begin{table}[ht!]
\centering
\begin{scriptsize}
\begin{tabular}{|l|l|}
\hline
\textbf{Parameter}                        & \textbf{Value} \\ \hline
\multicolumn{2}{|c|}{\textbf{[PAGE\_PARSER]}} \\ \hline
RUN\_LAYOUT\_PARSER                       & yes                  \\ \hline
RUN\_LINE\_CROPPER                        & yes                  \\ \hline
RUN\_OCR                                  & yes                  \\ \hline
RUN\_DECODER                              & no                   \\ \hline
\multicolumn{2}{|c|}{\textbf{[LAYOUT\_PARSER\_1]}} \\ \hline
METHOD                                    & LAYOUT\_CNN          \\ \hline
DETECT\_LINES                             & yes                  \\ \hline
DETECT\_REGIONS                           & yes                  \\ \hline
MERGE\_LINES                              & no                   \\ \hline
ADJUST\_HEIGHTS                           & no                   \\ \hline
MODEL\_PATH                               & ./ParseNet\_296000.pt \\ \hline
MAX\_MEGAPIXELS                           & 5                    \\ \hline
USE\_CPU                                  & no                   \\ \hline
DOWNSAMPLE                                & 5                    \\ \hline
DETECTION\_THRESHOLD                      & 0.2                  \\ \hline
LINE\_END\_WEIGHT                         & 1.0                  \\ \hline
VERTICAL\_LINE\_CONNECTION\_RANGE         & 3                    \\ \hline
SMOOTH\_LINE\_PREDICTIONS                 & no                   \\ \hline
\multicolumn{2}{|c|}{\textbf{[LAYOUT\_PARSER\_2]}} \\ \hline
METHOD                                    & REGION\_SORTER\_SMART \\ \hline
\multicolumn{2}{|c|}{\textbf{[LINE\_CROPPER]}} \\ \hline
INTERP                                    & 2                    \\ \hline
LINE\_SCALE                               & 1.0                  \\ \hline
LINE\_HEIGHT                              & 50                   \\ \hline
%\multicolumn{2}{|c|}{\textbf{[OCR]}} \\ \hline
%OCR\_JSON                                  & ./ocr\_engine.json    \\ \hline
%USE\_CPU                                  & no                   \\ \hline
%\multicolumn{2}{|c|}{\textbf{Other Parameters}} \\ \hline
%exported                                  & true                 \\ \hline
%checkpoint                                & ./OCR\_350000.pt     \\ \hline
%line\_px\_height                           & 50                   \\ \hline
%line\_vertical\_scale                      & 1.25                 \\ \hline
%net\_name                                  & VGG\_LSTM\_B64\_L17\_S4\_CB4 \\ \hline
%characters                                & \texttt{[511 unique characters]}       \\ \hline
\end{tabular}
\end{scriptsize}
\caption{Configuration Parameters for ParseNet}
\label{tab:paramsParseNet}
\end{table}

%\todo[inline]{Este parrafo habría que reescribirlo, no se entiende que se trata de contar.}
%%Where regions and baseline text are detected by applying specific configurations such as downsampling (DOWNSAMPLE=5),  maximum size of the input image (MAX\_MEGAPIXELS=5), setting detection thresholds (DETECTION\_THRESHOLD=0.2), and adjusting line connection ranges (VERTICAL\_LINE\_CONNECTION\_RANGE=3).

With respect to the environment configuration, it must be noted that the ParseNet component for layout analysis is a pretrained model with a configuration for all experiments according to the configuration settings described in Table~\ref{tab:paramsParseNet}. In contrast, the OCR component is based on a MAE-ViT model that must be trained specifically for each dataset. In order to train the models, it is necessary to count on page line images and the corresponding ground truth texts. Table~\ref{tab:linesForOCRtraining} indicates the number of line images that have been used for training, validation and testing in each dataset. It must be noted that in the case of the Molino dataset testing is directly performed with full text pages to evaluate the feasibility of the proposed pipeline to process the full page documents. In addition, Appendix~\ref{app:molino} explains the process for curating the page line images of the Molino dataset and the corresponding groundtruth text.

%So, the dataset used for this project consists of a total of 13,762 line text image, divided into two subsets: 12,385 images for training and 1,377 images for validation. This split ensures that the model has access to a sufficiently and diverse training set while reserving a dedicated portion for evaluating generalization performance. The training data was curated to capture a wide range of variation in the target domain, and the validation set was selected to reflect similar diversity, enabling reliable performance assessment during development. In addition, we have also selected some pages for testing 10 images from 705 pages of Miguel del Molino \cite{molino1585}. This collection contains old Latin characters that modern tools often fail to recognise or convert correctly. 

\begin{table}[h!] %% placement specifier
\centering %% Center alignment
\begin{scriptsize}
\begin{tabular}{|l | l | l| l |} %% Four columns (left, center, right, center)
\hline %% Horizontal line for table header
 \textbf{Dataset} & \textbf{Training images} &  \textbf{Validation images} &  \textbf{Testing images}\\  
\hline
Rodrigo & 9,000  & 1,000 & 5,010    \\  \hline
Bentham & 9,198  & 1,415 & 860  \\  \hline
Molino  & 12,385 &  1,377 & 10 pages (*) \\  \hline
\end{tabular}
\end{scriptsize}
\caption{Page line images for the MAE-ViT models in each dataset. Notes: (*) testing is directly performed with full text pages in the case of the Molino dataset.}
\label{tab:linesForOCRtraining}
\end{table}

\subsection{Transcription of handwritten text}
\label{sec:handwritten}

The first experiments with handwritten texts were performed with the Rodrigo dataset, which contains 107 distinct character classes, representing a diverse and historically nuanced alphabet. In addition, it must be clarified that the elements of this dataset are directly page line images. Therefore, as there is no need to recognize the layout or to reconsctuct the final output, these experiments are focused on testing separately the OCR compoment described in Section \ref{MAE}, i.e. the masked Autoencoders (MAE) with a Vision Transformer (ViT).

We trained the MAE-ViT model with different batch sizes. First, the model was trained with a batch size of 128 for testing and a smaller batch size of 8 for additional evaluation. Specifically, we employed a weight decay of 0.5, a mask ratio of 0.4, attention mask ratio of 0.1, a maximum span length of 8, input image size of 512×64, a projection dimension of 8, and a dilated erosion module with a maximum kernel size of 2 and one iteration. Training was carried out over 100,000 iterations with no post-processing applied. Under these settings, our model achieved a Character Error Rate (CER) of 1.30\% and a Word Error Rate (WER) of 6.97\% when tested with a batch size of 128. With a smaller batch size of 8, the model achieved CER of 1.68\% and WER of 9.27\%. As presented in Table~\ref{tab:model_comparisonRodrigo}, these results are competitive with existing state-of-the-art models such as the ones proposed by Granell et al. \cite{granell2018transcription} (a deep model based on a Convolutional Recurrent Neural Network -- CRNN) and the one proposed by Chammas et al. \cite{chammas2021fine} (a deeper CRNN backbone with Temporal Dropout -- CRNN + TD). Notably, our model reaches these levels of accuracy without any post-processing, emphasizing the strength of its core architecture and training approach. These results validate the robustness of the model and its adaptability across historical handwritten of varying complexity like Rodrigo dataset.

\begin{table}[h]
\centering
\begin{scriptsize}
\begin{tabular}{|l|c|c|}
\hline
\textbf{Model} & \textbf{CER (\%)} & \textbf{WER (\%)} \\
\hline 
Deep CRNN \cite{granell2018transcription}  & 3.00 & 14.00 \\
Deep CRNN + TD \cite{chammas2021fine}  & 2.43 & 12.31 \\
\textbf{FP-THD} with batch size 8 & 1.68 & 9.27 \\
\textbf{FP-THD} with batch size 128 & \textbf{1.30} & \textbf{6.97} \\
\hline
\end{tabular}
\end{scriptsize}
\caption{Comparison of CER and WER values with test data from the Rodrigo dataset.}
\label{tab:model_comparisonRodrigo}
\end{table}

We followed the experiments with handwritten texts with the Bentham dataset. 
Again, as this dataset consists of page line images, these experiments are focused on testing the performance of the MAE-ViT model.
The model was trained using a batch size of 64 for training and 16 for validation, a weight decay of 0.5, a mask ratio of 0.4, and an attention mask ratio of 0.1. The model utilizes dilated erosion with a maximum kernel size of 2 and one iteration, along with a maximum span length of 8, image input size of 512×64, and a projection dimension of 8. With a training regime of 100,000 iterations and 94 target character classes. Despite the increasing reliance on complex post-processing techniques in state-of-the-art handwritten text recognition, our model achieves a CER of 4.46\% and a WER of 7.68\% without the use of any post-processing or external language modeling. As showed in table \ref{tab:model_comparisonBentham}, when compared to recent models such as the Handwritten Text Recognition with Joint Attention Network and Knowledge Distillation (HTR-JAND) \cite{hamdan2024htr}, which reports a CER of 2.02\% with the aid of sophisticated post-processing techniques like T5-based contextual correction, our model stands as a compelling and efficient alternative. These results validate the effectiveness of our core system and provide a solid foundation for further improvements via lightweight decoding enhancements or attention-based refinements. 

\begin{table}[h]
\centering
\begin{scriptsize}
\begin{tabular}{|l|c|c|}
\hline
\textbf{Model} & \textbf{CER (\%)} & \textbf{WER (\%)} \\
\hline 
HTR-JAND \cite{hamdan2024htr}  & 2.02 & 4.23 \\
\textbf{FP-THD} & \textbf{4.46} & \textbf{7.68} \\
\hline
\end{tabular}
\end{scriptsize}
\caption{Comparison of CER and WER values with test data from the Bentham dataset.}
\label{tab:model_comparisonBentham}
\end{table}

% This performance highlights the strength and generalization capability of our architecture, 
% demonstrates strong robustness and competitive performance even in the absence of any post-correction

%%%%%%   comments:   We could explain more explicitly that 2 of the experiments are just focused on the tuning of models for HTR.

%In these two models, our focus is specifically on the task of handwritten text recognition (HTR). 

%To validate the robustness of MAE-ViT model \cite{li2025htr}, we trained and tested it on new datasets not used in the original work. The results demonstrate that the model produces strong performance even when applied to different data, confirming its generalization capabilities. 

It must be noted that Rodrigo and Bentham are not the same datasets as the ones used for the experiments in the original work of MAE-ViT model \cite{li2025htr}, which employed LAM \cite{lam}, IAM \cite{iam} and READ2016 \cite{read2016}. Therefore, our results demonstrate that the model produces strong performance even when applied to different data for training and testing, confirming its generalization capabilities.

%\textcolor{green}{The sentence means that we use a different dataset that are used in the original work: it can be replaced by this : We tested them on new datasets not used in the original study}

\subsection{Transcription of printed text}
\label{sec:printed}

%We develop a pipeline to transcribe full-page processing instead of a text line image.

The experiments with historical printed texts were performed with the Molino dataset. Table~\ref{tab:paramsMAE} summarizes the training configuration of the MAE-ViT model, based on a Vision Transformer (ViT) architecture. The model was trained with a maximum learning rate of 0.001, using a batch size of 64 for training and 8 for validation. To encourage robustness and efficient learning, 40\% of the input is masked during training, and an attention mask ratio of 10\% is applied, with a maximum attention span length of 8. The input images are resized to 512×64 pixels, and the projection size is set to 8. For the morphological augmentation stage, dilation and erosion operations are applied using a maximum kernel size of 2 for 1 iteration. Random sampling is controlled with a probability factor of 0.5, and the regularization is managed using a weight decay factor of 0.5 and an alpha parameter of 1. The model is trained for a total of 100,000 iterations. We experimented with an additional data augmentation involving random rotations of up to ±45 degrees, but this did not lead to improved performance. As a result, we decided to keep the original set of data augmentations described earlier, which already provided good results.
%We also experimented with an additional data augmentation by randomly rotating the input images within a range of ±45 degrees. However, this approach did not yield satisfactory results, and the model's performance decreased. Therefore, we concluded that this type of augmentation is not particularly useful for our task.

\begin{table}[h!]
\centering
\begin{scriptsize}
\begin{tabular}{|c|c|}
\hline
\textbf{Parameter}              & \textbf{Value}           \\ \hline
 
Maximum learning rate                &   \texttt{1e-3}                    \\ \hline
Training batch size        &   \texttt{64}                          \\ \hline
Validation batch size            &   \texttt{8}                          \\ \hline
Weight decay factor        &   \texttt{0.5}                           \\ \hline
Mask ratio for the input    &   \texttt{0.4}                    \\ \hline
Attention mask ratio     &   \texttt{0.1}                        \\ \hline
Maximum span length for attention     &   \texttt{8}            \\ \hline
Image size for input    &   \texttt{512 x 64}         \\ \hline
Projection size        &    \texttt{8}                               \\ \hline
Maximum kernel size for dilation erosion  &    \texttt{2}       \\ \hline
Number of iterations for dilation erosion  &  \texttt{1}      \\ \hline
 Probability factor for random sampling    & \texttt{0.5}      \\ \hline
Alpha parameter               &    \texttt{1}                              \\ \hline
 Total number of iterations    & \texttt{100000}  \\ \hline
\end{tabular}
\end{scriptsize}
\caption{Training configuration of MAE-ViT for the Molino dataset}
\label{tab:paramsMAE}
\end{table}

%TABLE 4: Parameters Used in the Experiment

After 100,000 training iterations, it achieved a CER of 1.43\% and a WER of 5.39\% on the validation set, demonstrating strong recognition performance and good generalization.
To evaluate the performance of our text recognition pipeline, we compared it against two existing transcription methods:

\begin{itemize}
\item The first one is the Pero-OCR engine \cite{kohut2021ts}, a publicly available system developed by Brno University of Technology. Pero-OCR is primarily designed for layout analysis and optical character recognition of both printed and handwritten European documents. It is particularly tailored to process low-quality Czech newspapers digitized from microfilms but has shown strong generalization capabilities across various printed materials and languages. The latest OCR models and APIs are accessible through the official website and GitHub repository. \footnote{\url{https://github.com/DCGM/pero-ocr/tree/master}}

\item The second baseline is the transcription provided by the Biblioteca Virtual del Patrimonio Bibliográfico (BVPB) \cite{molino1585}, which serves as an official digital edition of the historical document under study. According to the ALTO XML metadata, this transcription was automatically generated using the ABBY Recognition Server which is a proprietary OCR software known for its structured text recognition use in digital heritage projects. 
%The processing date indicates the content was generated in 2019, representing a high-quality, institutionally endorsed OCR output. 
\end{itemize}

\begin{table}[h!]
\centering
\begin{scriptsize}
\begin{tabular}{| l | c | c |}
\hline
\textbf{} & \textbf{CER (\%) } & \textbf{WER (\%)} \\
\hline
BVPB \cite{molino1585}          & 0.3379          & 0.6835          \\
Pero-OCR \cite{kohut2021ts}     & 0.0242          & 0.2106          \\
\textbf{FP-THD}    & \textbf{0.0178} & \textbf{0.0450} \\
\hline
\end{tabular}
\end{scriptsize}
\caption{Comparison of CER and WER values with 10 test pages from the Molino dataset.}
\label{tab:cer_wer_comparison}
\end{table}

Table \ref{tab:cer_wer_comparison} compares the performance of the three models on a transcription task using CER and WER over 10 pages of Miguel del Molino's book \cite{molino1585}. Our model achieves the best results, with the lowest CER (0.0178) and WER (0.0450), outperforming both Pero-OCR and ABBY transcriptions. Our model shows a substantial improvement in word-level accuracy, reducing the WER by nearly 80\% compared to Pero-OCR and over 90\% compared to BVPB. After a qualitative evaluation of the transcription results, it can be observed that our model preserves better tildes and other sensitive diacritics, which play a critical role with texts written in medieval Latin.

\section{Discussion}
\label{sec:discussion}

Apart from describing a pipeline system for parsing pages, it must be acknowledged that our approach efficiently handles different font types and mixed-language content. To evaluate our approach we have tested it with various datasets demonstrating that the MAE-ViT OCR component effectively processes different types of text, including handwritten and printed documents. In addition, we wanted to avoid the use of commercial models by developing an efficient and free tool that facilitates the transcription without the need of applying any post-processing steps to correct the output.

However, our pipeline has also some limitations. For instance, in the case of the Molino dataset we have observed that marginal notes are not always transcribed correctly, and short lines with only one word can also be missed. 
Nevertheless, our system handles properly the main content in the two columns of the page, preserving important features of Latin medieval texts like abbreviations, tildes, and other special characters.

\section{Conclusions}
\label{sec:conclusions}

%%This paper has presented a pipeline for recognizing text in historical documents, providing a complete solution for full-page transcription. In particular, tested for the transcription of historical printed texts in medieval Latin (with a set of random pages extracted from the book a well known civil law expert of XVI century), we achieve a CER of 1.7\% and a WER of 4.5\%, outperforming other existing transcriptions. The results with handwritten texts are also competitive with respect to other existing solutions, achieving a CER of 1.3\% on the Rodrigo dataset and 4.46\% on Bentham dataset. This demonstrates a strong performance, particularly on the Rodrigo dataset.

%%In the future, this pipeline will be integrated into a larger platform to support the analysis of civil law documents \cite{neji2024first}. The next planned research tasks in this project will be the automatic expansion of Latin abbreviations and the automatic translation from Latin to Spanish using Large Language Models. Therefore, obtaining a high accuracy in the initial Latin transcription is essential for the success in the following tasks.

This paper has proposed a pipeline, called FP-THD, for the transcription of historical printed texts in medieval Latin. We use an existing  layout analysis model followed by an extended masked autoencoder and vision transformer to transcribe image text that contains different types of characters. In addition, we annotate and construct our own dataset to train the model, as existing annotated data for historical pages in medieval Latin are limited. In addition, our work was not restricted only to the dataset we prepared: although the original focus of this pipeline was on printed documents, we also trained it on existing handwritten datasets. As a result, we obtained promising and sufficient outcomes, showing that our approach is not only a comparison, but also a noticeable improvement in performance. This highlights the superiority of the models and its capability to handle both printed and handwritten recognition. 

However, the approach also has limitations: since the annotated dataset primarily consisted of long text lines rather than single words, the model struggles with images containing only a single word, which restricts its performance in certain scenarios. Despite this, the results provide valuable insights for the field of historical Latin text recognition, showing that combining annotated data with existing resources can significantly improve recognition capabilities. 

In the future, we will focus on expanding the dataset to include isolated words, exploring domain adaptation techniques, and improving the model’s ability to generalize across different document layouts. In addition, this pipeline will be integrated into a larger platform to support the analysis of civil law documents \cite{neji2024first}. The next planned research tasks in this platform will be the automatic expansion of Latin abbreviations and the automatic translation from Latin to Spanish using Large Language Models. Therefore, the obtaining of high accuracy in the initial Latin transcription  is essential to ensure the success of the following subsequent tasks in our project.

%Text expansion and translation are planned as the next steps in this project, aiming to further reduce manual effort and enhance the usability of the transcriptions. This work establishes a solid basis for advancing the automatic processing of historical texts.

%The key contributions of this paper are summarized as follows:
%\begin{itemize}
 %\item 
%We present a pipeline for multipage text recognition as an innovative tool for recognizing text in historical documents.
%\item We facilitate the complete optical character recognition process for historical document for full-page transcription.
%\item 
%\end{itemize}

%This pipeline has achieved relevant performance results. Tested on the transcription of Molino dataset, we achieve character error recognition of 1.7\% and word error recognition of 4.5\% without expanded abbreviations. Additionally, our algorithm tested on the Rodrigo dataset improves the performance by 1.3\% character error recognition and 6.97\% word error recognition \todo[inline]{with respect to what}.

%The proposed pipeline for transcription has bee in integrated within a more ambitious platform to facilitate the analysis of civil law documentary heritage \cite{neji2024first}. The objective of this platform is to integrate open-source tools to assist with tasks such as transcription, translation, and information extraction, which usually require significant human resources when done manually. 

\section*{Funding}

This work was supported by the Aragon Regional Government (Spain) [grant numbers T59\_23R, PROY\_S11\_24].

%This paper is part of the R\&D projects T59\_23R and PROY\_S11\_24 supported by the Aragon Regional Government (Spain).

%Section text. See Subsection \ref{subsec1}.

%% Use \subsection commands to start a subsection.

%\appendix
%\section{Example Appendix Section}
%\label{app1}


%%       \cite{<label>} ==> [1]

%%
%Example citation, See \cite{lamport94}.


%\bibliographystyle{elsarticle-harv}

\bibliographystyle{elsarticle-num}
\bibliography{bibliography}% common bib file

\appendix
\begin{appendices}

\section{Curation of the Molino dataset}
\label{app:molino}

%\todo[inline]{If necessary, we can also incorporate some of the example figures created for deliverable E1.1. \textbf{Yes, maybe we can incorporate an examples of line text as showed on figure \ref{fig:examplesPages} if you agree with me i can keep it like that.}}

%\begin{table}[h!] %% placement specifier
%\centering %% Center alignment
%\begin{tabular}{|p{4cm} | c | c|} %% Four columns (left, center, right, center)
%\hline %% Horizontal line for table header
% \textbf{Book name} & \textbf{Train and Valid pages} &  \textbf{Train and Valid lines text}  \\  
%\hline
%Edition in 1667 of ``Fueros del Reino de Aragón'' \cite{fueros1667}  & 23  &    1277     \\  \hline
%Facsimile  1866 of ``Fueros del Reino de Aragón'' \cite{savall1866}  & 69  &    7263   \\  \hline
%Translation to Spanish ``Fueros del Reino de Aragón'' \cite{delgado1991}  & 15 &  2070     \\  \hline
%Edition in 1585 ``Fueros del Reino de Aragón'' \cite{molino1585} & 26 &   3152     \\  \hline
%Total & \textbf{134} & \textbf{ 13762}     \\  
%21 & 22 & 23 & 24 \\  
%\hline
%\end{tabular}
%\caption{Number of pages per book and the total of each split}
%\label{tab:molinoDataset}
%\end{table}

The preparation of the Molino dataset required the generation of the line images of every page and their corresponding texts. In order not to do it completely manually, we took profit of the existing pre-trained models of ParseNet (already described in Sections \ref{sec:relatedWork} and \ref{sec:layout}) and Pero-OCR.\footnote{https://github.com/DCGM/pero-ocr/tree/master} Pero-OCR is designed for OCR of both printed and handwritten text, which has been tailored to Czech newspapers. It delivers excellent performance on most European printed documents, including low-quality ones digitized from microfilms. 

Table \ref{tab:paramsOCR} describes all the parameters used on the transcription and generation of the text lines images with the correspondent text files. The Pero-OCR model  utilizes a configuration with a CNN for feature extraction, followed by LSTM layers for sequence modeling, trained with a batch size of 64, 17 layers, stride of 4, and 4 convolutional blocks and a line height of 50 pixels. The original line height was 40, but we adjusted it to be closer to the size required for the input of our MSE model. Using this approach (instead of training the Pero-OCR model from scratch), the amount of time and resources required for processing each page are reduced. 

Due to errors in the transcription of medieval Latin in Miguel del Molino's book using Pero-OCR, we had to correct manually part of the transcriptions, focusing only on the selected pages of this book. This manual correction was particularly important for handling Latin abbreviations, which often include tildes or special characters that require careful attention.

\begin{table}[ht!]
\centering
\begin{scriptsize}
\begin{tabular}{|l|l|}
\hline
\textbf{Parameter}                        & \textbf{Value} \\ \hline
\multicolumn{2}{|c|}{\textbf{[PAGE\_PARSER]}} \\ \hline
RUN\_LAYOUT\_PARSER                       & yes                  \\ \hline
RUN\_LINE\_CROPPER                        & yes                  \\ \hline
RUN\_OCR                                  & yes                  \\ \hline
RUN\_DECODER                              & no                   \\ \hline
\multicolumn{2}{|c|}{\textbf{[LAYOUT\_PARSER\_1]}} \\ \hline
METHOD                                    & LAYOUT\_CNN          \\ \hline
DETECT\_LINES                             & yes                  \\ \hline
DETECT\_REGIONS                           & yes                  \\ \hline
MERGE\_LINES                              & no                   \\ \hline
ADJUST\_HEIGHTS                           & no                   \\ \hline
MODEL\_PATH                               & ./ParseNet\_296000.pt \\ \hline
MAX\_MEGAPIXELS                           & 5                    \\ \hline
USE\_CPU                                  & no                   \\ \hline
DOWNSAMPLE                                & 5                    \\ \hline
DETECTION\_THRESHOLD                      & 0.2                  \\ \hline
LINE\_END\_WEIGHT                         & 1.0                  \\ \hline
VERTICAL\_LINE\_CONNECTION\_RANGE         & 3                    \\ \hline
SMOOTH\_LINE\_PREDICTIONS                 & no                   \\ \hline
\multicolumn{2}{|c|}{\textbf{[LAYOUT\_PARSER\_2]}} \\ \hline
METHOD                                    & REGION\_SORTER\_SMART \\ \hline
\multicolumn{2}{|c|}{\textbf{[LINE\_CROPPER]}} \\ \hline
INTERP                                    & 2                    \\ \hline
LINE\_SCALE                               & 1.0                  \\ \hline
LINE\_HEIGHT                              & 50                   \\ \hline
\multicolumn{2}{|c|}{\textbf{[OCR]}} \\ \hline
OCR\_JSON                                  & ./ocr\_engine.json    \\ \hline
USE\_CPU                                  & no                   \\ \hline
\multicolumn{2}{|c|}{\textbf{Other Parameters}} \\ \hline
exported                                  & true                 \\ \hline
checkpoint                                & ./OCR\_350000.pt     \\ \hline
line\_px\_height                           & 50                   \\ \hline
line\_vertical\_scale                      & 1.25                 \\ \hline
net\_name                                  & VGG\_LSTM\_B64\_L17\_S4\_CB4 \\ \hline
characters                                & \texttt{[511 unique characters]}       \\ \hline
\end{tabular}
\end{scriptsize}
\caption{Configuration Parameters for Layout and OCR Models}
\label{tab:paramsOCR}
\end{table}

\end{appendices}

%\end{thebibliography}
\end{document}